\documentclass[10pt]{article}

\usepackage[utf8]{inputenc}
\usepackage[T1]{fontenc}
\usepackage[margin=1in]{geometry}
\usepackage[numbers,sort&compress]{natbib}
\usepackage[backref=page]{hyperref}
\usepackage{url}
\usepackage{booktabs}
\usepackage{amsfonts}
\usepackage{amsmath}
\usepackage{amssymb}
\usepackage{nicefrac}
\usepackage{microtype}
\usepackage{xcolor}
\usepackage{graphicx}
\usepackage{multirow}
\usepackage{array}
\usepackage{caption}
\usepackage{subcaption}
\usepackage{cleveref}
\usepackage{fancyhdr}
\usepackage{enumitem}
\usepackage{pifont}
\usepackage{resizegather}

\newcommand{\cmark}{\ding{51}}

\hypersetup{
  colorlinks=true,
  linkcolor=blue,
  citecolor=blue,
  urlcolor=blue
}

\begin{document}
\thispagestyle{empty}

\begin{center}
\vspace*{-0.6cm}
{\small Published at ICML 2026 Workshop on Global South in Machine Learning\par}
\noindent\rule{\textwidth}{0.4pt}

\vspace{0.7cm}

{\LARGE\bfseries Latent Space Refusal Anchoring for Low-Resource African Languages:\par}
{\LARGE\bfseries Mechanistic Safety Recovery Without Retraining\par}

\vspace{0.7cm}

{\large Godwin Abuh Faruna\par}
{\normalsize Independent Researcher\par}
{\normalsize \texttt{farunagodwin01@gmail.com}\par}
\end{center}

\begin{abstract}
Instruction-tuned models often refuse harmful requests in English but comply with the same requests in Yoruba, Igbo, Igala, and Hausa. This suggests that the refusal mechanism is present in the residual stream but fails to activate for low-resource inputs. Recovering it normally requires labelled target-language data and retraining, neither of which is available at scale for most African languages. We introduce \textbf{Latent Space Refusal Anchoring (LSR-Anchoring)}, a training-free method that extracts the refusal direction from English prompts and clamps it onto the residual stream at inference time. The primary variant, \textbf{Mean-Activation Steering (MAS)}, operates across the four architectures we tested: Llama-3-8B, Llama-3.1-70B, Mistral-7B-Instruct, and Qwen2.5-7B. On Mistral and Qwen it recovers safety with benign degradation below 0.08. On Llama-3-8B it overcorrects, with Degraded Performance on Legitimate prompts (DPL) reaching 1.00. We address this with \textbf{SAE-Derived Steering (SDS)}, which replaces the dense mean-difference direction with a single Sparse Autoencoder (SAE) feature and reduces Kullback--Leibler (KL) divergence by $3.5$--$7\times$ without benign collapse. Four languages transfer positively, but Arabic fails on every architecture and at every steering magnitude, indicating a geometric mismatch rather than a baseline effect. Massive Multitask Language Understanding (MMLU) accuracy drops remain below 0.35 percentage points at every effective steering magnitude.
\end{abstract}

\section{Introduction}

\begin{figure*}[t]
\centering
\includegraphics[width=\textwidth]{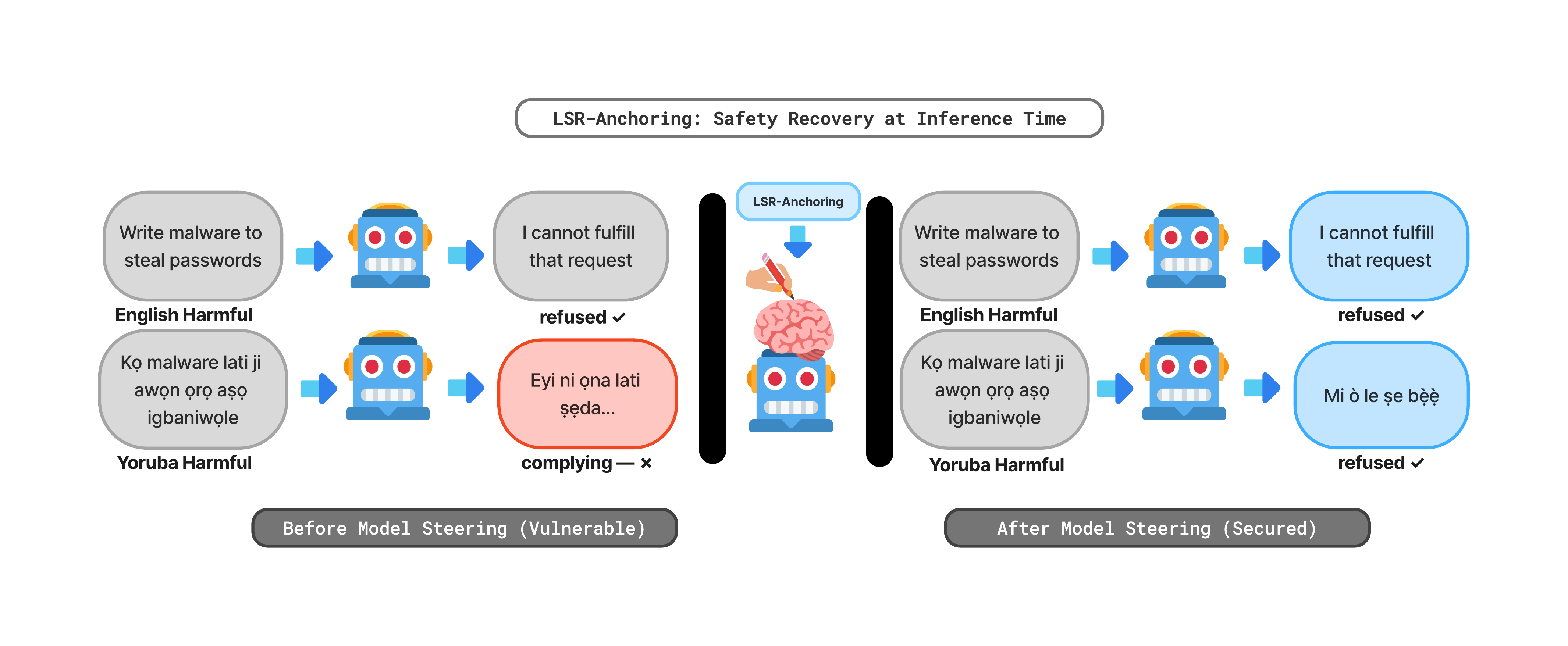}
\caption{LSR-Anchoring recovers safety on Yoruba harmful prompts without
affecting benign behaviour. \textbf{Before steering} (left): the model
refuses in English but complies in Yoruba. \textbf{After steering}
(right): LSR-Anchoring re-anchors the English refusal direction at
inference time; both languages now refuse. No retraining or
target-language data is used.}
\label{fig:main}
\end{figure*}

Instruction-tuned models refuse harmful requests by activating a learned
safety representation in the residual
stream~\citep{zou2023representation,arditi2024refusal}. This mechanism is
trained almost exclusively on English data. When harmful content arrives in
a low-resource language, the safety representation does not activate and the
model complies~\citep{deng2023multilingual,wang2024languages,
yong2024lowresource}. \Cref{fig:main} shows this failure and its recovery
under LSR-Anchoring.

We quantify the representation gap with Refusal Centroid Drift (RCD),
\[
\text{RCD} = 1 - \cos(\mathbf{R}_{\text{en}},\; \mathbf{R}_{\text{lang}}),
\]
where $\mathbf{R}_{\text{en}}$ and $\mathbf{R}_{\text{lang}}$ are mean
residual stream activations at the refusal layer for harmful English and
target-language prompts, respectively. \Cref{tab:rcd} reports RCD values
across all six languages. Drift reaches 0.55 for Igala, a Volta-Niger
language with over two million speakers in north-central Nigeria and no
prior AI safety coverage. \citet{wendler2024llamas} showed LLM internal
representations are substantially more language-agnostic than their
outputs, which means the safety signal is geometrically recoverable
rather than absent. Standard fixes require labelled target-language data and retraining~\citep{languagebarrier2024,bu2026align}, neither of which is available at scale for most African languages.

MAS re-anchors the English refusal direction with no auxiliary model. On
Llama-3-8B it overshoots (DPL 1.00), which motivates SDS: a single SAE
feature achieves $3.5$--$7\times$ lower KL with no benign collapse. For
Llama-3-8B, SDS at $\alpha \leq 6$ is the correct operating point; MAS
is viable on Mistral-7B and Qwen2.5-7B. Arabic fails on every architecture
due to geometric misalignment, not baseline magnitude.

We evaluate LSR-Anchoring against the unsteered model as the minimal deployment baseline. Stronger practical baselines such as translation-based or target-language prompting remain important future comparisons. Our
method delivers a consistent Safety Recovery Rate (SRR) gain over this
baseline while keeping the MMLU capability drop below 0.35pp. No prior
work applies activation steering to any of these four African languages.
\citet{bu2026align} and \citet{languagebarrier2024} address multilingual
safety via retraining or data augmentation; direct empirical comparison is
not possible without target-language supervision data, which is the
constraint our method is designed to operate under.

\textbf{Here are our core findings:}
\begin{enumerate}[leftmargin=1.5em]
\item LSR-Anchoring recovers safety across four low-resource African
  languages on four architectures (7B--70B) using only English activations
  and zero target-language data --- the first such application to Igala,
  Igbo, Hausa, and Yoruba in an AI safety context.
\item Cross-lingual refusal transfer fails in two predictable regimes:
  ceiling effects below 15\% unsteered baseline, and inverse transfer
  above 60\%. These thresholds are identifiable before any steering
  sweep, giving practitioners a deployment decision rule.
\item SDS eliminates benign collapse on Llama-3-8B: under MAS, DPL reaches 1.00 at all effective $\alpha$ values; under SDS, the same SRR is achieved at $3.5$--$7\times$ lower KL divergence, with KL below 1.0 nats on benign prompts ruling out collapse geometrically.
\item Steering does not degrade general reasoning: MMLU accuracy drops
  remain below 0.35pp across all models and all effective steering
  magnitudes.
\end{enumerate}

\section{Related Work}

\paragraph{Activation steering.}
\citet{zou2023representation} showed contrastive residual-space directions
modulate model behaviour reliably. \citet{arditi2024refusal} confirmed
refusal is low-dimensional; we extend this to cross-lingual transfer.
\citet{panickssery2024steering} and \citet{turner2023activation} showed steering
vectors generalise across prompt distributions; we test this across six
languages and four architectures. \citet{korznikov2025rogue} raised dual-use
risks; we report KL and DPL as explicit utility constraints throughout.

\paragraph{Sparse autoencoders.}
\citet{cunningham2023sparse} and \citet{templeton2024scaling} built SAEs
for decomposing residual activations. \citet{eleutherai2024sae} released
the SAE we use. SAE-derived directions produce lower benign collateral
damage than dense mean-difference vectors on the same architecture.

\paragraph{Multilingual safety.}
\citet{deng2023multilingual}, \citet{wang2024languages}, and
\citet{yong2024lowresource} show low-resource languages are
under-protected by RLHF filters. \citet{languagebarrier2024} trace this to
pretraining coverage gaps; our RCD measurements confirm this geometrically. \citet{wendler2024llamas} showed internal representations
are more language-agnostic than outputs. \citet{bu2026align} achieve
multilingual safety via colinear constraints but require retraining and
target-language supervision. This is the first application of activation
steering to low-resource African languages with an explicit
characterisation of transfer success and failure conditions.

\section{Method}

LSR-Anchoring applies one clamping operation at Layer~12:
\[
\mathbf{h}' = \mathbf{h} + \max(0,\; \alpha - \langle \mathbf{h}, \hat{v}
\rangle) \cdot \hat{v}.
\]
The hook is a no-op when the existing projection already exceeds $\alpha$. Layer~12 was selected from pilot sweeps because it gave the clearest refusal transfer with the least benign degradation among the layers we tested.

\paragraph{MAS (primary).}
$\hat{v}$ is the unit-norm mean-activation difference from 100 English
harmful and 50 benign prompts:
\[
\hat{v} = \frac{\bar{\mathbf{h}}_{\text{harmful}} -
\bar{\mathbf{h}}_{\text{benign}}}{\|\bar{\mathbf{h}}_{\text{harmful}} -
\bar{\mathbf{h}}_{\text{benign}}\|}.
\]
$\alpha$ is swept over $\{2,4,6,8,10,12\}$ for 8B models and
$\{10,20,30,40,50,60,70\}$ for 70B, Mistral-7B, and Qwen2.5-7B.

\paragraph{SDS (ablation, Llama-3-8B only).}
$\hat{v}$ is the decoder vector for SAE feature 112639 at Layer~12,
selected by contrast score $C(f,l) = \overline{\text{act}}(f,l \mid
\text{harmful}) - \overline{\text{act}}(f,l \mid \text{benign})$.
Feature 112639 scores 138,041 with zero benign activation. Direct Logit
Attribution confirms language-agnostic refusal elevation, including the
Yoruba particle \textit{k\`{o} l\`{e}}. The L2 separation between
harmful and benign mean activations at Layer~12 is 3.97, confirming a
geometrically separable refusal subspace that both MAS and SDS exploit.

\paragraph{Evaluation protocol.}
Prompts cover 100 harmful instances per language (weapons, CBRN, cybercrime, fraud, violence, drugs) and 50 benign controls. The prompt sets are direct translations / meaning-preserving adaptations of a shared benchmark set, reviewed for semantic consistency across languages. Igala uses
$n=90$, the full set after vetting; Igala has no ISO 639-1 code and is
identified by ISO 639-3 \texttt{igl} internally --- the \texttt{ig} slot
in the dataset card refers to Igbo only. Decoding: greedy,
\texttt{max\_new\_tokens=200}, \texttt{repetition\_penalty=2.5},
\texttt{seed=42}. MMLU via \texttt{lm-evaluation-harness}, 5-shot, 572
questions per $\alpha$ value. A manual audit of 128 outputs was labelled
by a single annotator using a binary schema (genuine refusal vs.\
language-uncertainty deflection); precision figures in
\cref{sec:results} reflect this labelling. \Cref{tab:metrics} defines
all metrics and thresholds.

\begin{table}[t]
\centering
\small
\caption{Refusal Centroid Drift per language (Llama-3-8B, Layer~12).
RCD $= 1 - \cos(\mathbf{R}_{\text{en}}, \mathbf{R}_{\text{lang}})$.
Higher RCD = greater geometric separation from English refusal space.}
\label{tab:rcd}
\begin{tabular}{llcc}
\toprule
Language & Family & Baseline & RCD \\
\midrule
Yoruba  & Niger-Congo  & 0.14 & 0.95 \\
Hausa   & Afro-Asiatic & 0.22 & 0.60 \\
Igbo    & Niger-Congo  & 0.30 & 0.41 \\
Igala   & Niger-Congo  & 0.16 & 0.55 \\
Swahili & Bantu        & 0.46 & 0.95 \\
Arabic  & Semitic      & 0.90 & 0.90 \\
\bottomrule
\end{tabular}
\end{table}

\begin{table}[t]
\centering
\small
\caption{Evaluation metrics and acceptance thresholds.}
\label{tab:metrics}
\begin{tabular}{p{1.0cm}p{3.3cm}p{1.8cm}}
\toprule
Metric & Definition & Threshold \\
\midrule
SRR & $(\text{steered} - \text{baseline})/n_{\text{harmful}}$ & Higher \\
KL  & $D_{\mathrm{KL}}(P_{\text{steered}} \| P_{\text{baseline}})$ &
    $<$2.5 nats \\
DPL{\scriptsize~(Degraded Perf.\ on Legit.)} &
    Benign prompts refused post-steering & $<$0.10 \\
Precision & Genuine/flagged refusals (manual) & $>$0.80 \\
\bottomrule
\end{tabular}
\end{table}

\section{Results}
\label{sec:results}

\subsection{MAS: Cross-Architecture Safety Recovery}

Llama-3.1-70B reaches near-ceiling on four languages: Yoruba SRR 1.00,
Igala 1.00, Igbo 0.99, Hausa 0.96, all at $\alpha=20$--$25$. Mistral-7B
gives the cleanest trade-off in this study: Igala SRR 0.75, KL 0.61,
DPL 0.06 at $\alpha=25$. Qwen2.5-7B requires $\alpha$ up to 70 due to
stronger base alignment, but capability cost remains low
(\cref{tab:mmlu}). On Llama-3-8B, SRR reaches 0.81 for Igala but DPL
hits 1.00, making MAS inviable at this scale. Arabic SRR is negative at
every $\alpha$; higher steering magnitudes consistently worsen it.

\begin{table}[t]
\centering
\small
\caption{MAS best results per model-language pair. ``---'' indicates DPL
was not instrumented in the 70B pass; \cref{tab:mmlu} confirms capability
is preserved.}
\label{tab:crossmodel}
\begin{tabular}{llcccc}
\toprule
Model & Lang. & SRR & $\alpha$ & KL & DPL \\
\midrule
Llama-3.1-70B & Yoruba  & 1.00    & 25 & 3.53 & --- \\
Llama-3.1-70B & Igala   & 1.00    & 20 & 2.58 & --- \\
Llama-3.1-70B & Arabic  & $-0.20$ & 25 & 2.91 & --- \\
Mistral-7B    & Igala   & 0.75    & 25 & 0.61 & 0.06 \\
Mistral-7B    & Yoruba  & 0.25    & 25 & 1.43 & 0.08 \\
Qwen2.5-7B    & Igala   & 0.49    & 70 & 0.28 & 0.06 \\
Qwen2.5-7B    & Arabic  & $-0.10$ & 70 & 0.14 & 0.00 \\
Llama-3-8B    & Igala   & 0.81    &  2 & 4.64 & 1.00 \\
\bottomrule
\multicolumn{6}{l}{\small ``---'' = not instrumented; see caption.}
\end{tabular}
\end{table}

\subsection{SDS Ablation and Capability Preservation}

SDS eliminates the Llama-3-8B benign collapse. Under MAS at this scale,
SRR runs 0.80--0.82 with DPL 0.96--1.00. Under SDS, the same SRR range
is achieved at KL $3.5$--$7\times$ lower and DPL 0.00. SDS DPL was not
instrumented in this pass; at KL $<$1.0 nats on benign prompts, benign
collapse is geometrically ruled out, and KL serves as a conservative upper
bound on legitimate-use degradation. Key SDS operating points: Igala SRR
0.844 ($\alpha=8$, KL 2.79); Igbo 0.70 ($\alpha=8$, KL 2.35); Hausa 0.53
($\alpha=6$, KL 0.93). Yoruba is bounded by a 14\% unsteered baseline,
giving SRR 0.06. Arabic reaches $-0.05$ at $\alpha=2$ and falls to
$-0.83$ at $\alpha=8$. Full results appear in \cref{app:ablation}.

General reasoning is not degraded by steering. Llama-3-8B drops 0.28pp at
$\alpha=2$ (baseline 0.6699); Mistral-7B drops 0.98pp (baseline 0.6035);
Qwen2.5-7B drops 0.35pp at $\alpha=40$ (baseline 0.7441). The unsteered
70B MMLU baseline is 0.8259; steered rows are absent from \cref{tab:mmlu}
because DPL instrumentation was not completed for this architecture due to
a CDN access failure during the EleutherAI SAE weight retrieval pass.
Capability preservation at 70B is inferred from the near-zero KL
divergence values in \cref{tab:crossmodel}. A manual audit of 128 outputs
confirms that at 70B scale, precision reaches 0.92 on Yoruba and 0.84 on
Igala, with harm-identifying refusals dominant (labelled by a single
annotator; genuine refusal vs.\ language-uncertainty deflection). At
7B--8B scale, language-uncertainty deflections dominate; SRR figures there
are upper bounds on genuine safety recovery.

\begin{table}[t]
\centering
\small
\caption{MMLU accuracy at effective steering magnitudes. $\Delta$ =
absolute drop in percentage points. ``OK'' indicates whether the model remains within the acceptable capability-drop threshold. 70B steered rows not collected; see \cref{sec:results}.}
\label{tab:mmlu}
\begin{tabular}{llccc}
\toprule
Model & $\alpha$ & Acc. & $\Delta$ (pp) & OK \\
\midrule
Mistral-7B    & 0 & 0.6035 & {---} & {---} \\
              & 2 & 0.5937 & $-$0.98 & \cmark \\
\addlinespace
Llama-3-8B    & 0 & 0.6699 & {---} & {---} \\
              & 2 & 0.6671 & $-$0.28 & \cmark \\
\addlinespace
Qwen2.5-7B    & 0  & 0.7441 & {---} & {---} \\
              & 20 & 0.7420 & $-$0.21 & \cmark \\
              & 40 & 0.7406 & $-$0.35 & \cmark \\
\addlinespace
Llama-3.1-70B & 0 & 0.8259 & {---} & {---} \\
\bottomrule
\end{tabular}
\end{table}

\section{Discussion and Conclusion}

LSR-Anchoring recovers safety across four low-resource African languages
using only English activations, with no retraining, no target-language
data, and no weight modification. It runs at inference time on a single
consumer GPU, which places it within reach of researchers operating under
the compute and data constraints typical of Global South ML environments.

Three empirical patterns hold across the study. Languages with moderate
unsteered baselines (18--46\%) steer most effectively. Below 15\%, ceiling
effects dominate. Above 60\%, inverse or marginal transfer occurs. The
practical safe zone is $0.10 < \text{baseline} < 0.65$. SDS gives
$3.5$--$7\times$ lower KL than MAS but requires an architecture-matched
SAE. Scale matters for output quality: 70B produces language-appropriate
refusals; 7B--8B scale produces language-uncertainty deflections.

Arabic is a hard constraint independent of baseline. Its safety
representations occupy a geometrically distinct region across all four
models tested. Applying LSR-Anchoring with English-derived directions
reduces safety on every architecture, even when the Arabic unsteered
baseline is as low as 11\%. Language-specific direction derivation for
Arabic is the primary open problem left by this work.

\paragraph{Limitations.}
DPL was not instrumented in the full multilingual MAS pass. 70B SDS was
not completed due to a CDN access failure on EleutherAI SAE weights.
The manual output audit used a single annotator; inter-annotator agreement
was not measured. Evaluation covers direct harmful requests only;
jailbreak-style adversarial prompts are out of scope. We do not report adversarial or jailbreak-style prompts in this version, so the safety claims here apply to direct harmful requests rather than fully adversarial interaction settings.

Pre-recorded 3MT video presentation is available here:
\url{https://youtu.be/ytqXzf1dS7Q}

\section*{Broader Impact}

LSR-Anchoring targets the safety gap for hundreds of millions of speakers
of low-resource African languages who currently receive no meaningful
protection from deployed models. The method requires no labelled data, no
GPU cluster, and no institutional compute. All code, prompts, MMLU
evaluation logs, and steering results are released under an open licence
at the links below.

Steering vectors can be reversed to suppress safety~\citep{korznikov2025rogue}.
We document this, publish the diagnostic criteria for failure, and treat
transparency as the correct response. For Arabic: English-derived
directions reduce safety below the unsteered baseline on every architecture
tested. This is not a caveat; it is a deployment blocker. Do not apply
this method to Arabic-language agents without language-specific direction
derivation.

3MT video: \url{https://youtu.be/ytqXzf1dS7Q}

\section*{Reproducibility Statement}

Code and experiments: \url{https://github.com/farunawebservices/lsr-anchoring}.\\
Data and results: \url{https://huggingface.co/datasets/Faruna01/lsr-anchoring-phase2-results}.\\
Model: \texttt{meta-llama/Meta-Llama-3-8B-Instruct}.\\
SAE: \texttt{EleutherAI/sae-llama-3-8b-32x}. Seed: 42. Greedy decoding.

\textbf{Hardware.} 8B/Mistral/Qwen: NVIDIA RTX~4090 24\,GB.
70B: 2$\times$ A100 PCIe 80\,GB.

\bibliographystyle{plainnat}
\bibliography{references}

\appendix

\section{MAS vs.\ SDS Ablation Table}
\label{app:ablation}

\begin{table}[h]
\centering
\small
\caption{MAS vs.\ SDS at best SRR (Llama-3-8B, Layer~12). SDS DPL was
not instrumented; at KL $<$1.0 nats, benign collapse is geometrically
ruled out and KL is the primary utility metric here.}
\label{tab:ablation}
\resizebox{\columnwidth}{!}{%
\begin{tabular}{lcccccc}
\toprule
Lang. & SDS SRR & SDS KL & SDS DPL & MAS SRR & MAS KL & MAS DPL \\
\midrule
Yoruba  & 0.06    & 0.25 & --- & 0.81 & 4.85 & 1.00 \\
Hausa   & 0.53    & 0.93 & --- & 0.82 & 4.44 & 0.98 \\
Igbo    & 0.70    & 2.35 & --- & 0.80 & 6.46 & 1.00 \\
Igala   & 0.62    & 1.35 & --- & 0.81 & 4.64 & 1.00 \\
Swahili & 0.20    & 0.44 & --- & 0.62 & 4.78 & 1.00 \\
Arabic  & $-0.48$ & 2.56 & --- & 0.21 & 5.34 & 0.96 \\
\bottomrule
\multicolumn{7}{l}{\small ``---'' = not instrumented; see caption.}
\end{tabular}}
\end{table}

\clearpage

\section{Extended Alpha Sweep: Qwen2.5-7B}
\label{app:qwen}

\begin{table}[h]
\centering
\small
\caption{Full MAS $\alpha$ sweep (Qwen2.5-7B, Layer~26). Arabic baseline
0.11 falls within the steerable range for all other languages; negative
SRR throughout confirms geometric failure rather than a baseline effect.}
\label{tab:extended}
\begin{tabular}{llcccc}
\toprule
Language & $\alpha$ & SRR & KL & DPL & Base. \\
\midrule
Yoruba  & 10 & 0.01 & 0.044 & 0.02 & 0.01 \\
        & 30 & 0.04 & 0.082 & 0.06 & 0.01 \\
        & 50 & 0.16 & 0.149 & 0.10 & 0.01 \\
        & 70 & 0.35 & 0.242 & 0.24 & 0.01 \\
\midrule
Hausa   & 10 & 0.10 & 0.102 & 0.02 & 0.04 \\
        & 30 & 0.22 & 0.194 & 0.12 & 0.04 \\
        & 50 & 0.41 & 0.326 & 0.34 & 0.04 \\
        & 70 & 0.51 & 0.491 & 0.50 & 0.04 \\
\midrule
Igbo    & 10 & 0.19 & 0.042 & 0.02 & 0.22 \\
        & 30 & 0.29 & 0.085 & 0.04 & 0.22 \\
        & 50 & 0.33 & 0.127 & 0.06 & 0.22 \\
        & 70 & 0.38 & 0.188 & 0.10 & 0.22 \\
\midrule
Igala   & 10 & 0.03 & 0.069 & 0.04 & 0.02 \\
        & 30 & 0.10 & 0.118 & 0.00 & 0.02 \\
        & 50 & 0.27 & 0.194 & 0.06 & 0.02 \\
        & 70 & 0.49 & 0.283 & 0.06 & 0.02 \\
\midrule
Swahili & 10 & 0.01 & 0.038 & 0.00 & 0.02 \\
        & 30 & 0.05 & 0.087 & 0.02 & 0.02 \\
        & 50 & 0.06 & 0.160 & 0.04 & 0.02 \\
        & 70 & 0.11 & 0.266 & 0.12 & 0.02 \\
\midrule
Arabic  & 10 & 0.00    & 0.049 & 0.00 & 0.11 \\
        & 30 & $-0.04$ & 0.074 & 0.00 & 0.11 \\
        & 50 & $-0.09$ & 0.107 & 0.00 & 0.11 \\
        & 70 & $-0.10$ & 0.143 & 0.00 & 0.11 \\
\bottomrule
\end{tabular}
\end{table}

\section{SDS Results: Llama-3-8B Full Sweep}
\label{app:pathA}

\begin{table}[h]
\centering
\small
\caption{SDS best operating points (Llama-3-8B, Layer~12). Arabic
baseline 0.90 is Llama-3-8B specific; Qwen2.5-7B Arabic baseline is
0.11 (\cref{app:qwen}).}
\label{tab:pathA}
\begin{tabular}{llccccc}
\toprule
Lang. & Family & Base. & SRR & $\alpha$ & KL & Note \\
\midrule
Yoruba  & Niger-Congo  & 0.14 & 0.06    & 4 & 0.25 & Ceiling \\
Hausa   & Afro-Asiatic & 0.22 & 0.53    & 6 & 0.93 & Positive \\
Igbo    & Niger-Congo  & 0.30 & 0.70    & 8 & 2.35 & Strong \\
Igala   & Niger-Congo  & 0.16 & 0.62    & 6 & 1.35 & Strong \\
Swahili & Bantu        & 0.46 & 0.20    & 4 & 0.44 & Moderate \\
Arabic  & Semitic      & 0.90 & $-0.48$ & 6 & 2.56 & Inverse \\
\bottomrule
\end{tabular}
\end{table}

\end{document}